# Study of Humanoid Push Recovery Based on Experiments

Vijay Bhaskar Semwal and Prof. G.C. Nandi
Robotics and AI Lab,
IIIT Allahabad, U.P-211012 (India)
{vsemwal, gcnandi}@gmail.com

*Abstract*— Human can negotiate and recovers from Push up to certain extent. The push recovery capability grows with age (a child has poor push recovery than an adult) and it is based on learning. A wrestler, for example, has better push recovery than an ordinary man. However, the mechanism of reactive push recovery is not known to us. We tried to understand the human learning mechanism by conducting several experiments. The subjects for the experiments were selected both as right handed and left handed. Pushes were induced from the behind with close eyes to observe the motor action as well as with open eyes to observe learning based reactive behaviors. Important observations show that the left handed and right handed persons negotiate pushes differently (in opposite manner). The present research describes some details about the experiments and the analyses of the results mainly obtained from the joint angle variations (both for ankle and hip joints) as the manifestation of perturbation. After smoothening the captured data through higher order polynomials, we feed them to our model which was developed exploiting the physics of an inverted pendulum and configured it as a representative of the subjects in the Webot simulation framework available in our laboratory. In each cases the model also could recover from the push for the same rage of perturbation which proves the correctness of the model. Hence the model now can provide greater insight to push recovery mechanism and can be used for determining push recovery strategy for humanoid robots. The paper claimed the push recovery is software engineering problem rather than hardware.

I. INTRODUCTION

Push recovery is the behavior shown by any subject towards recovery from unexpected external push. Most of the existing humanoid robots are bipedal. In spite of the fact that biped robot is more problematic from control perspective. We try to design the biped robots because they have better adaptability, obstacle negotiation capability and they can climb stairs.

Since they will have to work in an environment made for human beings who are inherently bipedal, they need to acquire human like push recovery capability. In case of human the efficient push recovery is based on learning. This is the reason why a child has a poor push recovery capability than an adult person because the child is in lower strata of the learning [1]. It is observed particularly in human locomotion, when a person is in drunk , the central pattern generatoris temporarily unstable, Which caused for not able to generate the rhythmic gait cycle necessary for walking, even though all the sensory input perfectly alright[4].

In this paper we have described how the challenges can be addressed. For this in a preliminary state we have developed a humanoid motion capturing device (HMCD) indigenously as shown in Fig-1. With the help of this device we have gathered a large number of data through an experimental framework designed in the Robotics & Artificial Intelligence laboratory of IIIT-A. The persons with the different age group, height and weight have been subjected to a push and their recovery actions have been captured for both male and female with left or right handed in nature i.e. humans have preference for left and right legs analogous to left and right handedness. We contributed an important analysis in this paper by unification of bipedal model and strategies used for bipedal balance and more stable walk. We further calculated the Centre of Mass and Centre of Pressure through accelerometer and gyro with 6 DoF with interface kit arudino, which is an important observation in the development of decision surface depicting boundary whether successfully preventing fall or failing.

So as to make the push sudden, cause and direction of push should be invisible for the person experiencing push; we have closed the eyes of the subject before applying push. The data have been analyzed using a linear inverted pendulum model [2].The parameters of the model we have configured so that the LIPM represents a humanoid robot HOAP2 [3] available in our laboratory.

The entire paper has been arranged as follows:

In a first section we have revealed our approach with coupled dynamic equation of a walking humanoid robot in their stance phase. Subsequently, we have presented benefits, scope and challenges in present paper followed by a methodology. In the result and discussion section we smoothened out our raw data using cubic spline-interpolation [5] and compared our captured experimental data of various joints corresponding to push. Comparing the result with standard data we have identified the recovery zone. We have concluded that the push recovery not only depends on the person's height, weight or sex but also depends on whether a person is a left or right handed. These analyses are important for further investigation and implementation on a real biped humanoid robot which will be synergistically similar with the human push recovery control in terms of partitioning the control algorithm. Some other relevant and recent works in the area of humanoid push recovery can be obtained in [4, 6]

II. OUR APPROACH

Push recovery is the behavior shown by any subject towards recovery from unexpected external push. The control reverse torques can be computed for the joints in a bipedal humanoid using the following equations:

$$\tau = M(\theta)\ddot{\theta} + C(\theta,\dot{\theta}) + G(\theta) \quad (1)$$

For normal walking pattern, torques can be generated as in eqn.(1) which is sum of inertial torque (M), Centrifugal and

Coriolis torque(C) and gravitational torque (G) together with some frictional torques which has been neglected. When external force is applied on person, these ideal walking patterns of different joint torques get disturbed. To regain original torques, control torques required to be applied. The mass matrix $M(\theta)$ is calculated through mass property of body, which is configuration dependent [8][18].

The experimental data of human can be directly applied on our LIPM model either by using fuzzy data set or Neural Network (NN) training set. Here we have used both types of approaches using MATLAB tool box.

### A. Benefits, scopes and challenges

It completely replaces the older concept of push recovery. Unlike existing humanoid robot push recovery strategy, human does not require simultaneous control on each joint. In human, one joint acts as an active joint and play vital role in push recovery. This leads to a new concept of push recovery pattern generated by a person depends on whether that person is left handed or right handed. We can apply real data on LIPM. There are lots of challenges faced during this experiment like (a) difference between human and robot torso (b) the gait pattern taken from HMCD is not ideal gait pattern of human (c) available size of force sensor is only 14.5cm² (d) digital counter in Phidget kit having a range of 0 to 999 counts only.

### B. Methodology

HMCD, which is used to extract real joint angle data for our experimental framework. It is designed indigenously in our lab using two aluminum rods which are connected by a 100 kΩ potentiometer. As the force is applied from the back, the angle between rigid aluminum rods will change which will be reflected by potentiometer reading between 0 to 300 degree.

Force is applied in eight different ways i.e. open eyes with lunging(hand movement) static, open eyes with lunging dynamic, open eyes without lunging static, open eyes without lunging dynamic, closed eyes with lunging static, closed eyes with lunging dynamic, closed eyes without lunging static, closed eyes without lunging dynamic. It can be further processed the data for learning LIPM by fuzzy and ANN network using MATLAB tool box.

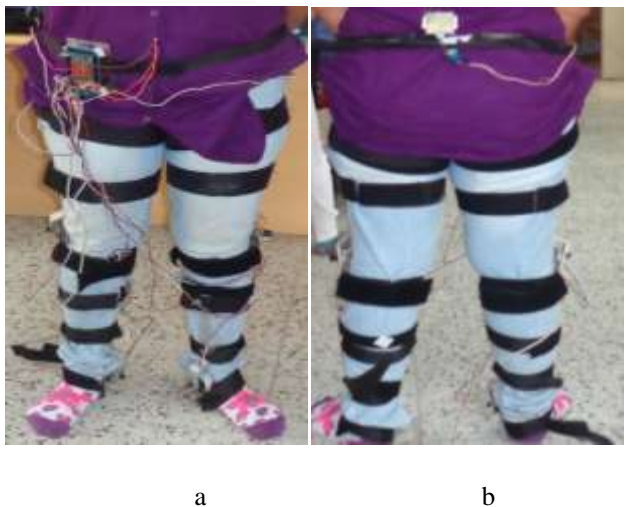

Fig.1. A subject wearing HMCD a). Front View with Accelerometer b). Back view with force sensor.

While gathering the data, zero correction is required. It has been done by getting $\Theta_o$ for all the initial position of the joint angles (hip, knee and ankle), and this $\Theta_o$ is subtracted from $\Theta$ (the joint angle when force is applied) which subsequently provides the change in joint angles at the instant of push.

All the potentiometer of HMCD is connected to Phidget interface kit (P/N 1018 8/8/8). It gives value in digital counts (within range 0 to 999) from which we can further convert it into angular values using the given formula [6].

$$Angle(Degree) = \left[(\theta - \theta_0) * \frac{300}{100}\right] \quad (2)$$

Force sensor (FSR 3105-Force sensing resistor) has been used with voltage divider (1121) on the back of person between last rib and L1 of spinal rod to measure applied force which is connected through the Phidget interface kit[9][10]. The data acquired from the force sensor is also in form of digital counts and then these data can be calculated in the units of force by using the given formula.

$$\text{Force (in Newton)} = f * (9.8/1000) \quad (3)$$

Where, f = force in digital counts

Force is applied from only one direction, due to limited area of force sensor, the range of FSR digital count is 1N to 100N and the sensing area of FSR 3105 is 14.5 cm². we used a wooden hammer like structure to give push forces. Image of FSR is given below [7].To increase FSR surface area used rectangular FSR surface.

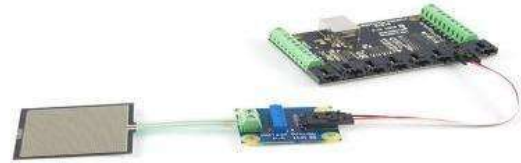

Fig.2. FSR with voltage divider connected in phidget kit.

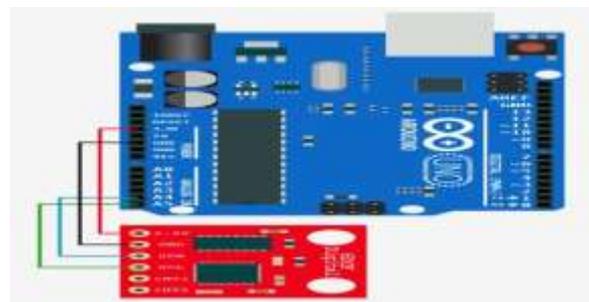

Fig.3. Ardunio interface Kit-. ITG3200/ADXL345

To measure the value of centre of mass for 6Dof we developed our own kit using IMU Digital Combo Board with ITG3200 gyro and ADXL345 accelerometer [14]. The IMU gives the orientation and acceleration info about center of mass.

1-ADXL345 accelerometer having range +/-2, 4, 8,16g, which is used to measure the CoM (Centre of mass) Acceleration[14].

2- ITG3200 gyro (range ±2000°/s) used to measure rotation [16].

### III. RESULTS AND DISCUSSION

Ideally a person have oscillatory motion in hip, having two humps in knee and two sharp humps like mountain in normal walking pattern without external force. The ideal graph is shown in figure 4[1].

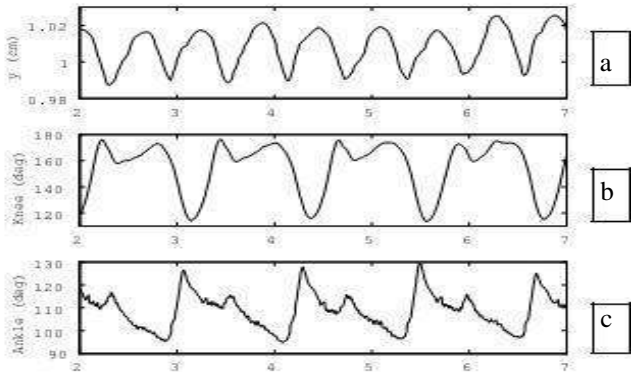

Fig.4. It shows an ideal gait pattern of human i.e. (a) hip, (b) knee and (c) ankle respectively.

In our discussion, graph of right and left handed person is in specific posture i.e. closed eye without lunging (hand movement in balancing) static. Figure no 4 shows the graph of right hip of right handed person. Here we are giving some push force to person on above mentioned location, horizontal axis as time duration and vertical axis as angle of joint in degree. In given graph we can see an oscillatory motion of hip, very close to ideal one. This can further be used in the analysis of crouch or abnormal subject. By studying we can improve gait cycle pattern of amputees and develop artificial limbs. The experimental data collected through in house developed HMCD device used in development of software based model[13] which is a step toward to proved our hypothesis [14], the push recovery can solve using software model [15][18].

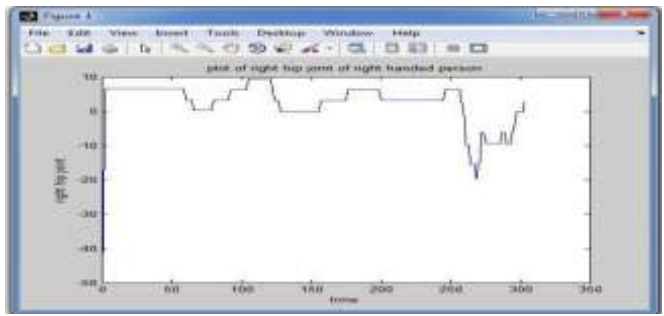

Fig.4.Graph of right hip of right handed person when push is applied.

The graph shown in figure no 5 is graph of left hip of right handed person we can see a deviation due to force and a person involve his left portion against push forces. It does not have continuous oscillatory motion in hip due to pressure on a left hip that sustain 1st force causes constant angle at joint then some movement in hip.

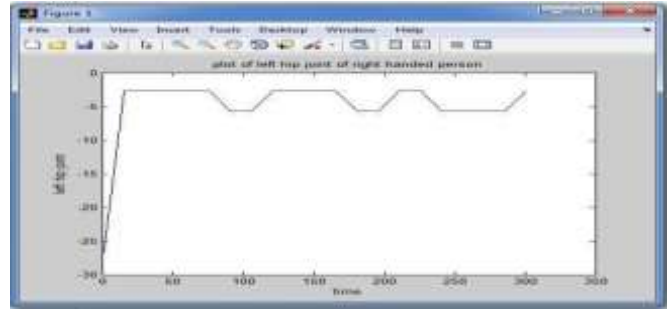

Fig.5.Graph of left hip of right handed person when push is applied.

From the above description of both the hip joint of right handed person as right hip is closed to ideal one and deviation in left hip from the ideal shows that left hip of right handed person is more active joint in push recovery against external force.

Figure no. 6 shows the Graph of right knee showing small disturbance due to balancing posture of a person. It seems like constant or stable during push recovery.

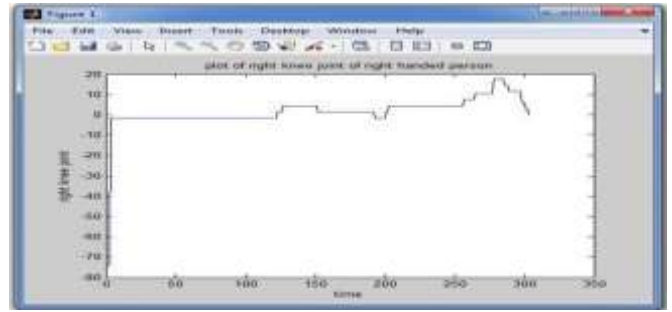

Fig.6.Graph of right knee of right handed person when push is applied.

Knee joint has maximum contribution in push recovery pattern generate by human in all three joint. In figure no.7 Graph of left knee shows very high angle deviation in knee joint. For right handed person its left portion is more active than right portion specially knee joint. These deviations in knee joint shows vital contribution of left knee in push recovery.

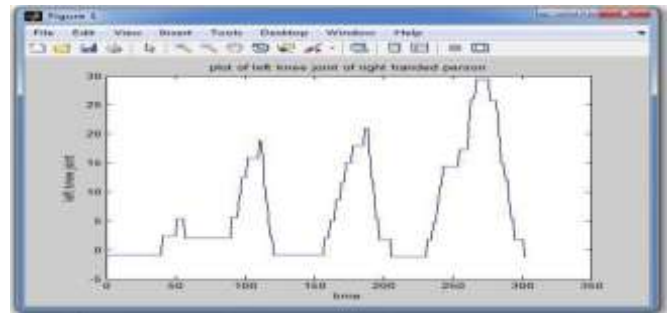

Fig.7.Graph of left knee of right handed person when push is applied.

Figure no. 8 given below shows the push recovery pattern right ankle of right handed person when push is applied.

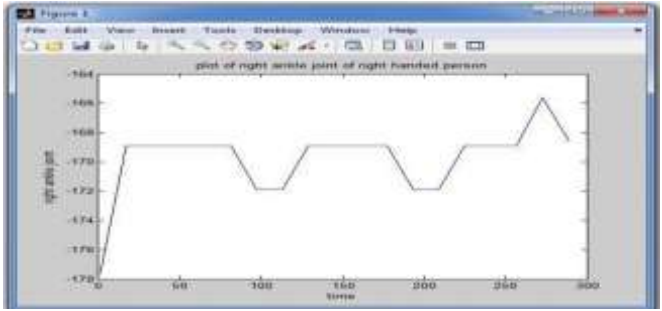

Fig.8 Graph of Right ankle of Right handed person.

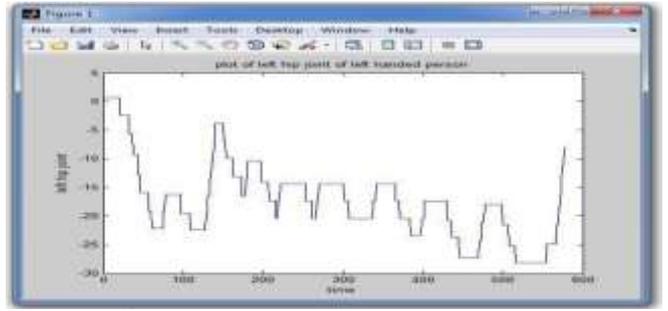

Fig.11.Graph of left hip of left handed person when push is applied

Figure no. 9 given below shows the recovery pattern of left ankle of right handed person when push is applied. It seems from both ankle joint that, ankle joint has less contribution in push recovery pattern of human. But in comparison of right ankle, left ankle is showing more deviation in angle in right handed person.

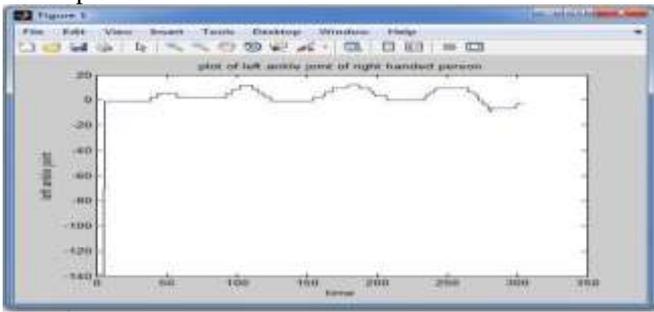

fig.9. Graph of Left ankle of Right handed person

Graph in figure no. 10 shows the recovery pattern of right hip angle and oscillation of left handed person. The graph is plotted between angles of hip joint with respect to time.

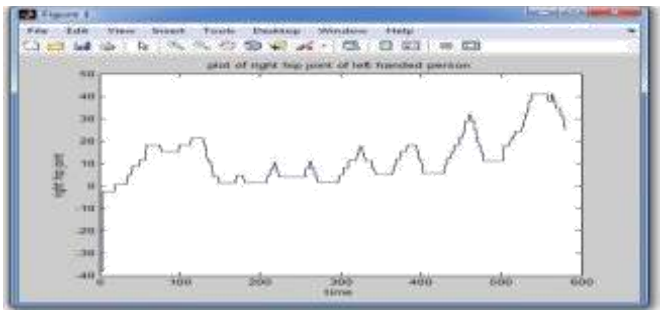

Fig.10.Graph of right hip of left handed person when push is applied

Graph in figure no.11 shows the pattern of left hip of left handed person generate when force is applied on torso. If we study the nature of both the graph 10 and 11 we can find that both graphs is mirror image of each other with phase difference of 180°.Graph of hip joint shows oscillatory motion and swing of the angle of hip joint.

Figure no. 12 shows the graph recovery pattern of right knee of left handed person. Right part of left handed person is more active specially knee. From the given graph we can see angle deviation of right joint and oscillatory motion shows change angle due to force to balance itself. Figure no. 13 shows the graph of left knee of left handed person when force is applied to subject to the torso. This graph shows little deviation in the angle of left knee of left handed person. Hence it shows that in recovery against push forces left knee of left handed person show small contribution. In figure no.14 shows the right ankle of left handed person. Graph given below shows very little deviation in the ankle joint. As we see in figure no. 15 graph of left ankle of left handed person have more deviation in comparison of left knee. We can see that knee of right portion it shows maximum deviation than other two joint.

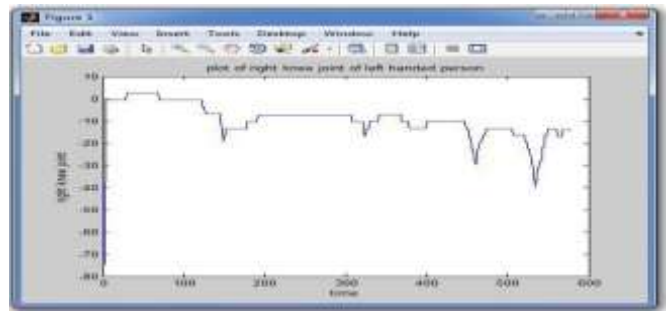

Fig.12.Graph of right knee of left handed person when push is applied

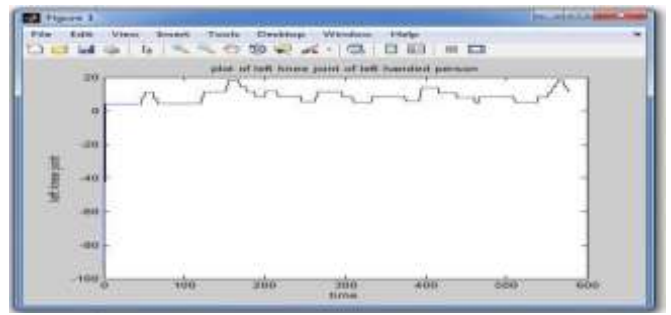

Fig13.Graph of left knee of left handed person when push is applied.

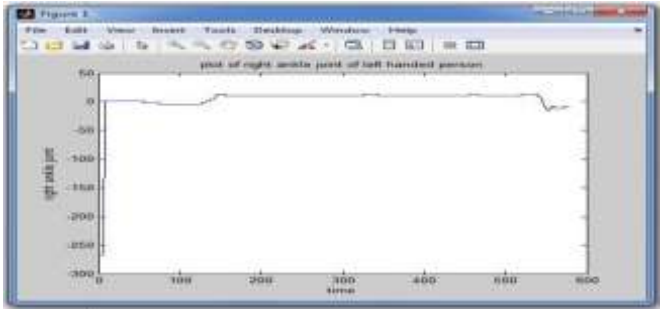

Fig.14 Graph of Right Ankle of left handed person.

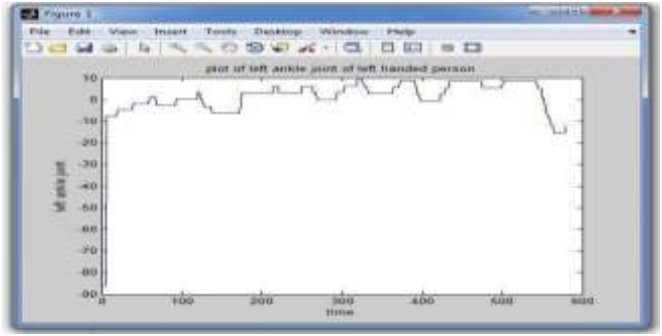

Fig.15 Graph of Left Ankle of left handed person.

As we compare graph of knee and ankle of left portion and right portion of left handed person we conclude that recovery pattern of joint angle of knee and ankle are inversely proportional to each other. If knee joint have more deviation then corresponding ankle joint have less deviation and vice versa.

The ideal and real phase plot of COM (Center of Mass) has shown in the figures given below. This graph tells about the stability of the walking pattern after applying the external force.

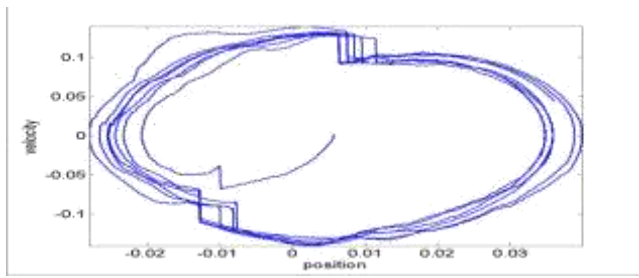

Fig16.Ideal phase plot of a COM

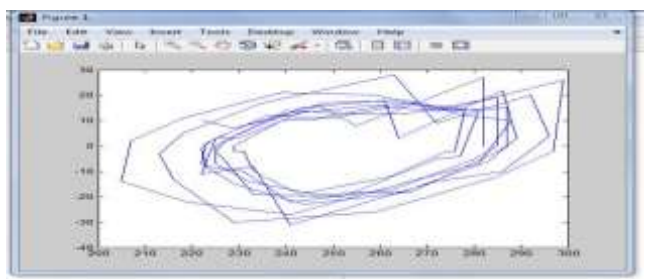

Fig17.Real phase plot of CoM

In the static mode[12][16][17], the phase plot has a decision boundary and if the phase plot lies below this decision boundary line then the person will[17] [8] able to recover from the external push and if the plot lies above this decision boundary then the person will not able to recover from the push.

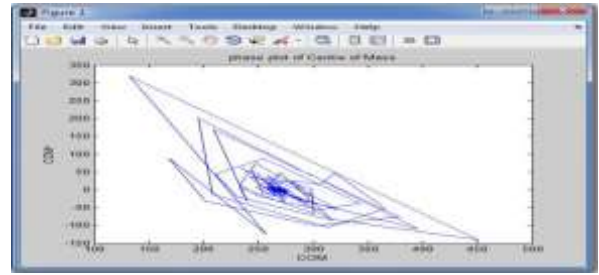

Fig18. Phase plate of CoM of human in a static mode.

The measurement of CoP (centre of pressure) revealed that with increase the magnitude of external force, the value of force at CoP on left foot is more than right foot, for a right handed person.

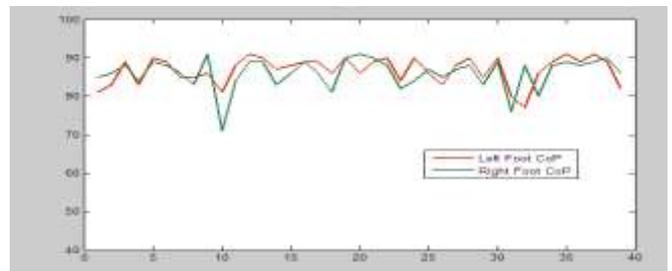

Fig.19. Phase plate of CoP of right handed person.

From the above comparison and study the pattern of six joint of right and left handed person we see that left portion of right handed person and right portion of left handed person is more active, i.e. for left handed person, his right leg joints are more active in push recovery pattern and vice versa. Vital role has played by knee in balancing.

CONCLUSION AND FUTURE WORK

The research shows some useful behaviour of human being towards push recovery. The strategy is different for left handed or right handed persons. The experimentally result using in house development proved that humans have preference for left and right legs analogous to left and right handedness. The knee joint is very active in maintaining balance towards push. It gives great insight to develop a reactive controller for a bipedal humanoid robot which we are going to develop subsequently. There is a coupling among the hip joints, knee joints during walking. We can say that push recovery is complex and non linear problem as it depends upon the parameters of the humans. In static mode, phase plot of centre of mass of a person tell us that the person is recovered from the push or fail. The result is an important contribution toward our hypothesis, the push recovery is a software engineering problem.